\documentclass[10pt,twocolumn,letterpaper]{article}

\usepackage{iccv}
\usepackage{times}
\usepackage{epsfig}
\usepackage{graphicx}
\usepackage{amsmath}
\usepackage{amssymb}
\usepackage{booktabs}
\usepackage[dvipsnames,table]{xcolor} 
\usepackage{xspace}
\usepackage{graphicx}
\usepackage{amssymb}
\usepackage{amsmath}
\usepackage{ dsfont }
\usepackage{multirow}
\usepackage{float}
\usepackage{lipsum}
\usepackage{nicefrac}
\usepackage{circledsteps}
\usepackage{enumitem}

\definecolor{lightgray}{gray}{0.95}
\definecolor{midgray}{gray}{0.55}
\definecolor{steelblue}{HTML}{4D82B7}
\definecolor{greenoo}{HTML}{16A03D}
\definecolor{blueoo}{HTML}{007AB0}

\newcommand{\errino}{\raisebox{0.25em}{\textcolor{blueoo}{\scriptsize{R}}}\xspace}
\newcommand{\jayino}{\raisebox{0.25em}{\textcolor{greenoo}{\scriptsize{JS}}}\xspace}
\newcommand{\result}[3]{\ensuremath{#1} \footnotesize{(\ensuremath{#3}})}
\newcommand{\resultj}[3]{\ensuremath{#1}\jayino \footnotesize{(\ensuremath{#3}})}
\newcommand{\resultr}[3]{\ensuremath{#1}\errino \footnotesize{(\ensuremath{#3}})}

\newcommand{\resultVAR}[2]{\ensuremath{#1}\scriptsize{$\pm$\ensuremath{#2}}}

\newcommand{\methodname}{Continual Learning via Equivariant Regularization\xspace}
\newcommand{\methnam}{CLER\xspace}
\newcommand{\classil}{oCIL\xspace}
\newcommand{\taskil}{oTIL\xspace}
\newcommand{\miniimagenet}{\textit{mini}ImageNet\xspace}

\usepackage{amsmath,pict2e}
\usepackage{gensymb}

\makeatother

\usepackage[pagebackref=true,breaklinks=true,letterpaper=true,colorlinks,bookmarks=false]{hyperref}

\iccvfinalcopy 

\def\iccvPaperID{0000} 

\ificcvfinal\fi

\begin{document}

\title{On the Effectiveness of Equivariant Regularization\\for Robust Online Continual Learning}

\author{Lorenzo Bonicelli$^1$ ~ Matteo Boschini$^1$ ~ Emanuele Frascaroli$^1$ ~ Angelo Porrello$^1$ ~ Matteo Pennisi$^2$\\Giovanni Bellitto$^2$ ~~ Simone Palazzo$^2$ ~~ Concetto Spampinato$^2$ ~~ Simone Calderara$^1$\vspace{0.5em}\\
$^1$AImageLab - University of Modena and Reggio Emilia \\
$^2$PeRCeiVe Lab - University of Catania\vspace{-0.75em}}

\maketitle
\ificcvfinal\thispagestyle{empty}\fi

\begin{abstract}
Humans can learn incrementally, whereas neural networks forget previously acquired information catastrophically. Continual Learning (CL) approaches seek to bridge this gap by facilitating the transfer of knowledge to both previous tasks (backward transfer) and future ones (forward transfer) during training.
Recent research has shown that self-supervision can produce versatile models that can generalize well to diverse downstream tasks. However, contrastive self-supervised learning (CSSL), a popular self-supervision technique, has limited effectiveness in online CL (OCL). OCL only permits one iteration of the input dataset, and CSSL's low sample efficiency hinders its use on the input data-stream.

In this work, we propose \textbf{\methodname (\methnam)}, an OCL approach that leverages \textit{equivariant} tasks for self-supervision, avoiding CSSL's limitations. Our method represents the first attempt at combining equivariant knowledge with CL and can be easily integrated with existing OCL methods. Extensive ablations shed light on how equivariant pretext tasks affect the network's information flow and its impact on CL dynamics.
\end{abstract}
\vspace{-1.45em}
%
\section{Introduction}
\label{sec:intro}
When dealing with non-stationary input distributions, Artificial Neural Networks (ANNs) show a bias towards the incoming training data and thus \textit{forget} previously acquired knowledge \textit{catastrophically}~\cite{mccloskey1989catastrophic}. Continual Learning (CL) is a rapidly growing area of machine learning that aims at designing approaches to counteract this effect~\cite{parisi2019continual,de2019continual}. Based on either parameter segregation~\cite{mallya2018packnet,serra2018overcoming}, regularization~\cite{kirkpatrick2017overcoming, li2017learning} or replay~\cite{robins1995catastrophic, buzzega2020dark, caccia2022new} -- CL methods allow machine learning systems to adapt constantly while remaining effective on old data. To assess the merits of these works, a plethora of experimental settings have been proposed in recent years; among those, we focus on the challenging Online CL (OCL) scenario~\cite{aljundi2019online,chaudhry2021using,caccia2022new} in light of its applicability to real-world problems: as it only allows a single pass on training data, it embodies the realistic assumption that an in-the-wild CL learner would hardly ever be exposed to the same input twice.
\begin{figure}[t]
    {
    \setlength{\tabcolsep}{0.1em}
    \centering
    \begin{tabular}{cc}
        \multicolumn{2}{c}{\includegraphics[width=\linewidth,trim={0 .8cm 0 .5cm},clip]{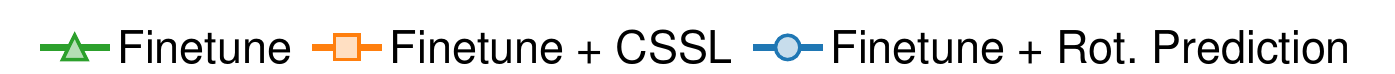}} \\
        \multicolumn{2}{c}{\includegraphics[width=\linewidth]{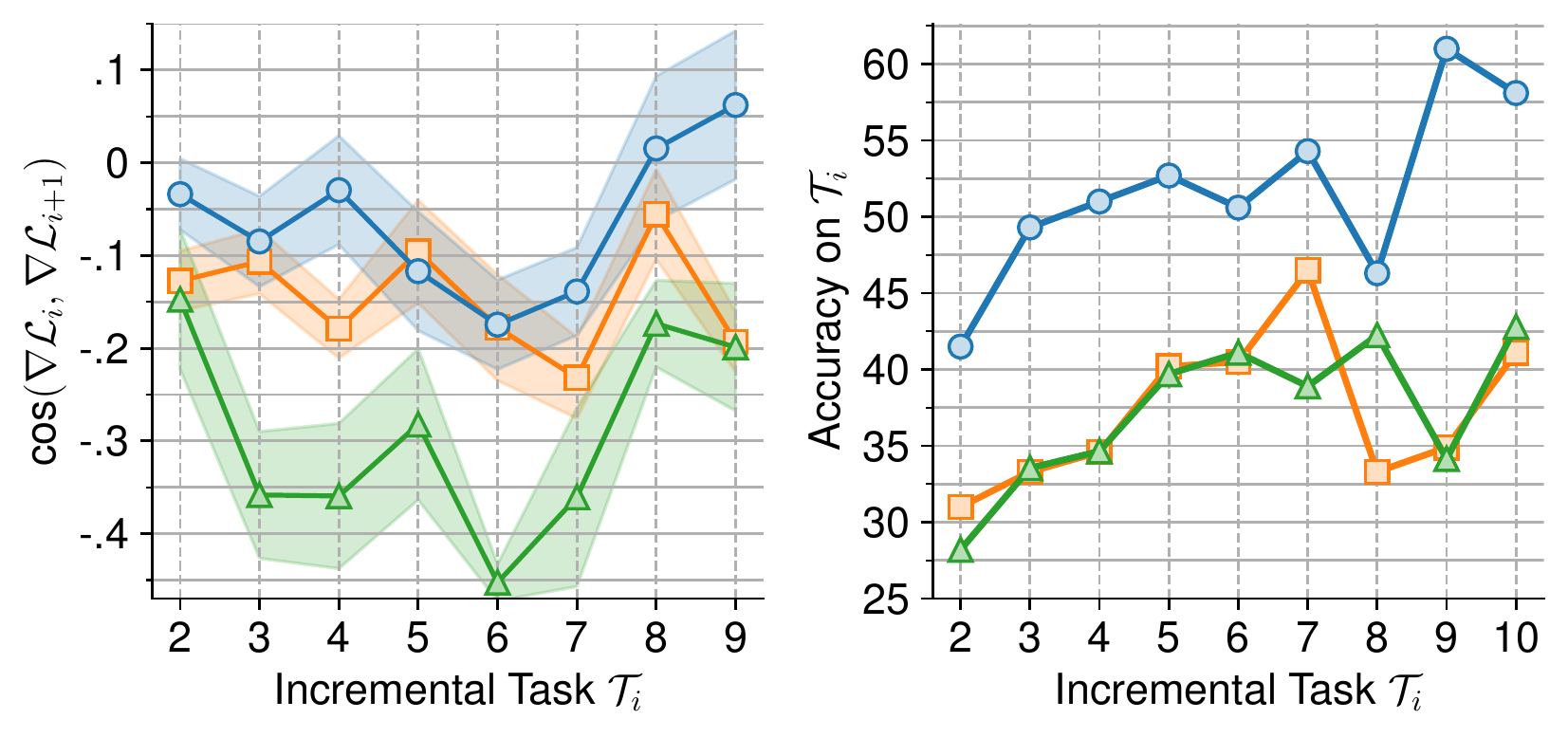}} \\
        \hspace{6.5em}(a) & \hspace{6.5em}(b) \\
    \end{tabular}
    }
    \caption{\textbf{Effects of SSL in OCL} (Seq.\ CIFAR-100) comparing a \textbf{Finetuning baseline} with no additional regularization (\textit{green}), with a \textbf{Contrastive SSL} auxiliary objective (\textit{orange}) and with an \textbf{Equivariant rotation prediction} pretext task (\textit{blue}). (a) Similarity between the gradients induced on the model by task $\mathcal{T}_i$ and $\mathcal{T}_{i+1}$ after training on $\mathcal{T}_i$. (b) Accuracy on task $\mathcal{T}_i$ after training on $\mathcal{T}_i$. Results are reported after a warm-up task (\textit{best in colors}).}
    \label{fig:sgd-test}
    \vspace{-0.8em}
\end{figure}

Motivated by the success of Contrastive Self-Supervised Learning (CSSL)~\cite{chen2020exploring,zbontar2021barlow,bardes2021vicreg}, several recent CL approaches pivot on self-supervised representation learning~\cite{pham2021dualnet,cha2021co2l,fini2022self,madaan2021rethinking}. Indeed, as self-supervised representations are generally acknowledged to be agnostic and easily transferable to diverse downstream tasks~\cite{chen2020simple}, their exploitation appears especially promising in the online scenario, where learning a shared representation across tasks is as important as the prevention of forgetting. Moreover, we argue that binding the incoming classes to general-purpose representations encourages the emergence of a horizontal and shareable knowledge base, that will be less subject to forgetting.

However, we reckon that the CSSL paradigm is not a silver bullet: indeed, contrastive methods are characterized by low \textit{sample efficiency} as their convergence requires large amounts of resources. As a result, CL methods need a higher number of training epochs when equipped with contrastive regularization~\cite{cha2021co2l}, which clashes with the constraints of OCL. Moreover, they usually focus their representation learning on a small memory buffer~\cite{pham2021dualnet}, which entails a high risk of overfitting~\cite{bonicelli2022effectiveness}.

This work addresses these limitations, revealing the benefits of \textit{equivariant} self-supervised tasks (\textit{e.g.}, rotation prediction, jigsaw puzzle, ...) for the OCL scenario. To provide an insight, Fig.~\ref{fig:sgd-test} considers a simple learner based on Finetuning (\textit{i.e.}, no counter-measure against forgetting) and reports its performance in the online scenario allowing only one epoch per task: in doing so, we compare the effects of the auxiliary objective based either on equivariant self-supervised learning (in this case, four-fold rotation prediction) or on Barlow Twins~\cite{zbontar2021barlow}, a recent CSSL-based approach that has also shown its merit in CL~\cite{pham2021dualnet}.
We observe that both representation learning tasks allow for a lower interference between features learned by SSL, as supported by the more favorable alignment of gradients between current and subsequent tasks (Fig.~\ref{fig:sgd-test}a). Surprisingly, Fig.~\ref{fig:sgd-test}b shows that only the rotation-aided model has a significant profit in terms of individual task accuracy for the CSSL-based objective. We conjecture that the limited amount of training steps in online CL is not sufficient for contrastive approaches (such as Barlow Twins) to produce effective representations for the downstream task. 

To address the aforementioned CSSL limitations in the OCL setting, we propose \textbf{\methodname (\methnam)}, a novel OCL regularizer built on top of equivariant pretext tasks -- to the best of our knowledge, this is the first attempt to exploit equivariant information in CL. We demonstrate that our proposal can be easily combined with existing state-of-the-art CL approaches, leading to a generalized improvement in performance. Through additional experiments, we highlight the structural and predictive properties conferred by \methnam and draw a detailed comparison with CSSL-based alternatives.
\section{Related Work}
\noindent\textbf{(Online) Continual Learning} is a field of machine learning that studies training over sequences of non-i.i.d.\ tasks, with the objective of retaining as much knowledge as possible from older tasks and mitigating catastrophic forgetting~\cite{mccloskey1989catastrophic}. The existing literature offers different techniques to tackle this problem: \textit{regularization-based}~\cite{kirkpatrick2017overcoming, li2017learning} methods are designed to control parameter updates in order to prevent disruptive modifications to features important for previous tasks; \textit{segregation-based}~\cite{mallya2018packnet,serra2018overcoming} approaches identify subsets of task-relevant parameters and prevent their alteration by combining parameter freezing, model expansion, and feature gating; \textit{replay-based}~\cite{robins1995catastrophic,riemer2018learning,buzzega2020dark,caccia2022new} methods store examples from the past in a memory buffer, with the objective of periodically refreshing older knowledge. Despite its simplicity, the latter approach is usually regarded as the most effective solution to date~\cite{farquhar2018towards,van2019three,chaudhry2019tiny}.

These methods are typically evaluated in a relaxed training setting, where the current task can be experienced over multiple epochs. In practical applications, this requirement is rarely satisfied; Online CL (OCL)~\cite{mai2022online,lopez2017gradient,aljundi2019gradient} is a challenging and realistic scenario that adds the condition that each sample of the stream can be seen only once. Works targeting OCL typically all belong to the \textit{replay-based} family~\cite{lopez2017gradient,chaudhry2019tiny}\footnote{All contemporary OCL works consider only replay approaches, due to their clear performance superiority over all alternatives~\cite{mai2022online,caccia2022new}.}. Among recent proposals, MIR~\cite{aljundi2019online} and GSS~\cite{aljundi2019gradient} propose enhanced replay sample selection procedures, ER-AML/ER-ACE~\cite{caccia2022new} encourage balance in learning by means of carefully designed loss functions, CoPE~\cite{de2021continual} learns by exploiting slowly evolving class summaries. 

\noindent\textbf{Self-Supervised Representation Learning in CL.}~Self-Supervised Learning aims at learning useful representations directly from the data, \textit{i.e.}, with no need for manual annotations. Recent SSL works show that these methods are able to learn strong representations that can reach or even outperform those of supervised learning~\cite{chen2020simple,chen2020exploring,zbontar2021barlow}. In the context of CL, SSL methods are typically trained to encourage the backbone network to be invariant to the given transformations~\cite{cha2021co2l,fini2022self,pham2021dualnet,madaan2021rethinking,kim2021continual}. Co\textsuperscript{2}L~\cite{cha2021co2l} learns the representations for new tasks with a modified supervised contrastive learning procedure~\cite{khosla2020supervised}, where current task samples are used as anchors and elements in the buffer are used as negative samples -- all this while preserving past knowledge through distillation. However, applying SSL methods in CL is not straightforward: SSL benefits from large batch sizes and require several training steps to converge~\cite{chen2020simple}; this represents a limit for Co\textsuperscript{2}L, as the number of negative samples is limited by the small buffer size.
DualNet~\cite{pham2021dualnet} decouples representation learning from the CL objective through two complementary networks: a \emph{slow net} exploits buffer samples to learn an overall representation, while a \emph{fast net} sequentially learns from the input stream, using the features from the slow net to guide the process. 

\noindent\textbf{Pretext Self-Supervised Learning and Rotations.}~Differently from CSSL,~\cite{gidaris2018unsupervised} employs a \textit{four-fold rotation} prediction pretext task to provide a powerful learning signal for representation learning. 
In~\cite{gidaris2019boosting}, the rotation pretext task is applied in the context of few-shot learning; similarly, \cite{dangovski2022equivariant} pairs rotation prediction to existing SSL methods, leading to a consistent performance improvement. Recently, the authors of~\cite{addepalli2022towards} investigated the role of invariance and equivariance in SSL, suggesting that some transformations (\textit{e.g.}, four-fold rotations, jigsaw puzzle) can be effective when employed to encourage equivariance, but can lead to disruptive effects when enforcing invariance.
\begin{figure*}[t]
    \centering
\includegraphics[width=0.98\textwidth,trim={2.02cm 5.21cm 8.1cm 1.19cm},clip]{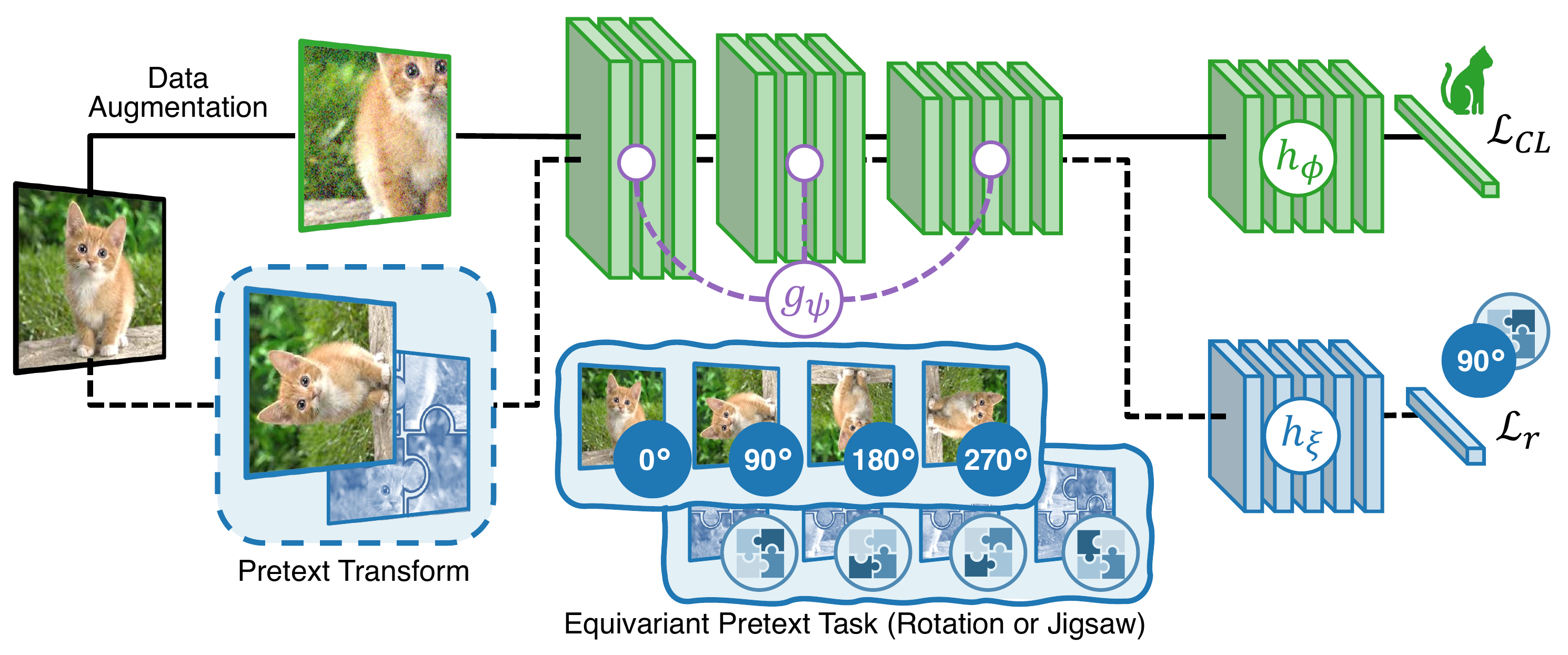}
    \caption{\textbf{Overview of \methnam}. Two versions of the input image are fed into the in-training model: \textit{i)} standard data augmentation is used to train the classification head (\textit{\textcolor{greenoo}{green}}); \textit{ii)} an equivariant transformation-based task (rotation, alternatively jigsaw) is used to train the pretext head (\textit{\textcolor{blueoo}{blue}}) (\textit{best in colors}).}
    
    \label{fig:model}
\end{figure*}
\section{Method}
\subsection{Online Continual Learning}
In Online Continual Learning (OCL)~\cite{aljundi2019gradient,chaudhry2021using}, a single DNN $f_\theta$ is trained on a sequence of classification tasks $\mathcal{T}_1,\dots,\mathcal{T}_T$. Each task consists of disjoint input and output distributions ($\mathcal{T}_i = (\mathcal{X}_i,\mathcal{Y}_i)$, with $\mathcal{Y}_i \cap \mathcal{Y}_j = \emptyset$ for $i\neq j$) and each example-label pair may only be shown to the model once. At task $\mathcal{T}_c$, CL aims at optimizing $f_\theta$ on all $T$ tasks, while only having access to data from $\mathcal{T}_c$ itself:
\begin{equation}
\label{eq:tr_obj}
    \mathcal{L} = \sum_{i=1}^{T} \mathcal{R}_i = \underbrace{\sum_{i=1}^{c-1} \mathcal{R}_i}_{\substack{\text{\Circled{1}}\\\text{data no longer}\\\text{available}}} + \underbrace{\mathcal{R}_c}_{\substack{\text{\Circled{2}}\\\text{data available}}} + \underbrace{\sum_{j=c+1}^{T} \mathcal{R}_j}_{\substack{\text{\Circled{3}}\\\text{data not yet}\\\text{available}}},
\end{equation}
where $\mathcal{R}_i = \mathbb{E}_{(x,y)\in\mathcal{T}_i}\big[\ell(f_\theta(x),y)\big]$ denotes the empirical risk associated with the data of task $\mathcal{T}_i$.

In Eq.~\ref{eq:tr_obj}, term \Circled{1} (stability) requires $f_\theta$ to maintain predictive efficacy on previously encountered data, whereas term \Circled{3} (plasticity) suggests that the model should prepare for fitting novel data distributions in later tasks. Only \Circled{2} can be directly pursued by training on data; instead, \Circled{1} and \Circled{3} are achieved by means of auxiliary loss terms. CL methods endeavor to balance the three terms, which are typically understood to interfere with one another~\cite{riemer2018learning,araujo2022entropy,lin2022towards}.

\subsection{OCL via Equivariant Regularization}
The objectives \Circled{1} and \Circled{3} from Eq.~\ref{eq:tr_obj} characterize the main challenges that come when designing a CL model. However, both can be addressed by learning a representation that can be shared across multiple tasks. To achieve this, we equip the online learner with an auxiliary SSL objective.
Works in current literature pursue this objective through CSSL loss terms~\cite{cha2021co2l,pham2021dualnet}; instead, we follow the insights presented in Sec.~\ref{sec:intro} and opt for an \textit{equivariant} pretext task~\cite{dangovski2022equivariant}, defined as follows.

Let $\mathcal{A} = \{\mathcal{A}_i\}_{i=1}^{K}$ be a family of input transforms $\mathcal{A}_i:\mathcal{X}\rightarrow\mathcal{X}$ (\textit{e.g.},\ rotations, jigsaw puzzle), we transform each input exemplar with a randomly chosen $\mathcal{A}_k$ and request the in-training model to recognize the transformation by predicting the correct label $k \in \mathcal{Y}_\mathcal{A} = \{1, \dots, K\}$. For this purpose, we rewrite $f_\theta$ as $h_\phi \circ g_\psi$, where $g_\psi$ is the early part of the network, devoted to the extraction of features, and $h_\phi$ encompasses the latter part of the model, including the final multi-layer classification head for the CL task. Subsequently, we introduce $h_\xi$: a separate sub-network following the same structure as $h_\phi$, finally projecting the representation $g_\psi(\cdot)$ on the set $\mathcal{Y}_\mathcal{A}$. 

We treat the choice of $\mathcal{A}$ as a hyperparameter. In our experiments, we explore two different kinds of transformations: the set of $4$ non-distorting image rotations $\{\operatorname{Rot}_{0\degree},\allowbreak\operatorname{Rot}_{90\degree},\allowbreak\operatorname{Rot}_{180\degree},\allowbreak\operatorname{Rot}_{270\degree}\}$~\cite{gidaris2019boosting,gidaris2018unsupervised}, and the $24$ permutations of patches produced by a $2\times 2$ jigsaw puzzle~\cite{noroozi2016unsupervised}.
The resulting approach, called \methnam, consists of a regularization term $\mathcal{L}_r$ that can be readily applied on a backbone network as shown in Fig.~\ref{fig:model}. Let $\mathbf{x}\in\mathbf{B}_\text{in}$ be a sample coming from the input batch, we define $\mathcal{L}_r$ as:
\begin{equation}
\label{eq:raider}
    \mathcal{L}_r = \lambda_r\cdot\mathop{\mathbb{E}}_{\substack{\mathbf{x}\sim\mathbf{B}_\text{in}\\k\sim\mathcal{Y}_\mathcal{A}}}\Bigg[  \operatorname{CE}\Big(h_\xi(g_\psi(\mathcal{A}_k(\mathbf{x}))), k\Big)\Bigg],
\end{equation}
where $\operatorname{CE}$ is the cross-entropy loss and $\lambda_r$ is a scalar hyper-parameter to control the strength of the regularization.
We highlight that the label space $\mathcal{Y}_\mathcal{A}$ of the pretext task remains constant over time. The objective of \methnam can hence be compared to classification problems where only the data-generating distribution is subject to changes (Domain-Incremental learning~\cite{van2019three}).

\smallskip
\noindent\textbf{Equivariance \& invariance}.~A function $f_\theta$ is said to be equivariant w.r.t.\ $\mathcal{A}$ if there exists a mapping $\mathcal{M}_\mathcal{A}$ such that:
\begin{equation}
\label{eq:pretext}
    f_\theta(T(\mathbf{x})) = \mathcal{M}_\mathcal{A}(f_\theta(\mathbf{x})),\quad\forall\mathbf{x}\in\mathcal{X}.
\end{equation}
While the learning objective in Eq.~\ref{eq:raider} promotes sensitivity to the chosen set of transformations, solving the CL task forces the model to become invariant w.r.t.\ employed data augmentations. To avoid overlapping between the two objectives, we compute Eq.~\ref{eq:raider} only on non-augmented inputs.
\section{Experiments}
\label{sec:exps}

\subsection{Experimental setting}

\noindent\textbf{Benchmarks.}~We build our OCL benchmarks by taking image classification datasets and splitting their classes equally into a series of disjoint tasks. In the online learning scenario, the learner will then experience each task \textbf{only once} (single epoch). For additional details regarding the experiments, we refer the reader to the supplementary material.
\begin{itemize}[noitemsep,leftmargin=*]
    \item \textbf{Seq.\ CIFAR-100}~\cite{zenke2017continual,rebuffi2017icarl,chaudhry2019tiny} is obtained by splitting the original $100$ classes of CIFAR-100~\cite{krizhevsky2009learning} into $10$ consecutive tasks. For each class, train and test sets include $500$ and $100$ $32\times 32$ RGB images respectively.
    \item \textbf{Seq.\ \miniimagenet}~\cite{chaudhry2019tiny,ebrahimi2020adversarial,derakhshani2021kernel} is a challenging dataset that includes a total of $100$ classes from the popular ImageNet dataset and a longer sequence of tasks. While the number of samples is the same as in Seq.\ CIFAR-100, images are resized to $84\times 84$ and split into $20$ $5$-way tasks. 
\end{itemize}

\noindent\textbf{Evaluation protocol.}~We primarily focus our evaluation on the online Class-Incremental (\classil) setting, where the model is asked to gradually learn to solve all tasks, with no information regarding the task identifier (Task-ID). Differently from the online Task-Incremental (\taskil) setting, where the task Task-ID is available during inference, \classil forces the learner to build a single-headed classifier. We present extensive results in both the \classil and \taskil settings.

\smallskip
\noindent\textbf{Baseline methods.}~We report the results of \methnam on a selection of current state-of-the-art (SOTA) methods viable for the \classil setting.
\begin{itemize}[noitemsep,leftmargin=*]
    \item \textbf{Experience Replay with Asymmetric Cross-Entropy (ER-ACE)}~\cite{caccia2022new}. Starting from the popular store-and-replay baseline (Experience Replay~\cite{ratcliff1990connectionist,robins1995catastrophic}), the authors propose an alteration aimed at preventing imbalances due to the simultaneous optimization of current and past data.
    \item \textbf{eXtended Dark Experience Replay (X-DER)}~\cite{boschini2022class} is a model that combines replay with self-distillation, while adopting careful design choices to harmonically blend predictive functions learned at different times.
    \item \textbf{Continual Prototype Evolution: Learning Online from Non-Stationary Data Streams (CoPE)}~\cite{de2021continual} proposes a classifier based on class prototypes, whose careful update scheme allows for learning incrementally while avoiding sudden disruptions in the latent space. %
    \item \textbf{DualNet}~\cite{pham2021dualnet} is a dual-backbone architecture decoupling the issue of incremental classification from the one of learning an overall transferable representation. The latter task is demanded to one of the backbones (\textit{slow learner}), trained with a CSSL loss term on i.i.d.\ data coming from the replay buffer; the other backbone (\textit{fast learner}) is instead tasked with fitting the CL tasks while taking advantage of the representations produced by the slow learner.
\end{itemize}
All models are trained for a single epoch with SGD, with a fixed batch size of $10$ both on the input stream and the replay buffer. We benchmark all models with two different sizes for the memory buffer: $500$ and $2000$ for Seq.\ CIFAR-100 and $2000$ and $8000$ for Seq.\ \miniimagenet. For these methods the input $\mathbf{B}_\text{in}$ in Eq.~\ref{eq:raider} is the concatenation of the images coming both from the stream and the buffer.
\begin{table*}[t] 
    \centering
    \rowcolors{2}{}{lightgray}
    \begin{tabular}{lcccc}
    \toprule
    \textbf{\classil} & \multicolumn{2}{c}{\textbf{Seq.\ CIFAR-100}} & \multicolumn{2}{c}{\textbf{Seq.\ \miniimagenet}} \\
    \midrule    
    Joint-offline & \multicolumn{2}{c}{\result{69.47}{0.20}{-}}                    & \multicolumn{2}{c}{\result{63.31}{0.47}{-}} \\
    Joint-online & \multicolumn{2}{c}{\result{23.14}{0.74}{-}}                    & \multicolumn{2}{c}{\result{10.68}{0.67}{-}} \\
    Finetune        & \multicolumn{2}{c}{\result{7.00}{0.15}{100}}    & \multicolumn{2}{c}{\result{3.21}{0.10}{100}}          \\
    \midrule
\rowcolor{white}\textbf{Buffer Size} & 500 & 2000 & 2000 & 8000 \\
    \midrule
    ER-ACE~\cite{caccia2022new}      & \result{20.17}{1.31}{38.75} & \result{26.95}{1.12}{23.69} & \result{15.03}{0.63}{35.01} & \result{16.07}{1.29}{37.94} \\
    ~+~\methnam & \resultj{\textbf{24.53}}{1.43}{\underline{33.76}} & \resultj{\textbf{30.89}}{1.54}{\underline{20.24}} & \resultr{\textbf{18.08}}{1.46}{\underline{32.53}} & \resultj{\textbf{18.43}}{0.95}{\underline{33.22}} \\
    X-DER~\cite{boschini2022class}   & \result{25.80}{0.98}{39.54} & \result{30.44}{0.80}{31.52} & \result{17.51}{1.08}{34.25} & \result{18.01}{1.54}{50.84} \\
    ~+~\methnam & \resultj{\textbf{29.35}}{0.72}{\underline{35.56}} & \resultj{\textbf{34.57}}{0.96}{\underline{29.71}} & \resultj{\textbf{21.26}}{0.91}{\underline{34.07}} & \resultj{\textbf{21.71}}{1.30}{\underline{34.76}} \\
    CoPE~\cite{de2021continual}      & \result{19.98}{1.49}{75.32} & \result{34.09}{2.12}{46.39} & \result{22.67}{1.32}{57.96} & \result{24.54}{2.67}{55.09} \\
    ~+~\methnam & \resultj{\textbf{26.15}}{1.16}{\underline{69.28}} & \resultj{\textbf{38.48}}{1.48}{\underline{45.50}} & \resultr{\textbf{25.91}}{1.67}{\underline{57.73}} & \resultr{\textbf{26.76}}{1.17}{\underline{52.69}} \\
    \midrule
    DualNet~\cite{pham2021dualnet}   & \result{11.09}{0.51}{92.42} & \result{19.93}{0.74}{73.44} & \result{16.21}{1.13}{80.35} & \result{25.33}{1.47}{59.60} \\
    ~+~\methnam & \resultr{\textbf{11.89}}{0.61}{\underline{89.97}} & \resultj{\textbf{20.88}}{0.90}{\underline{73.02}} & \resultr{\textbf{18.66}}{1.00}{\underline{72.74}} & \resultr{\textbf{30.90}}{0.86}{\underline{52.14}} \\
    \bottomrule
    \end{tabular}
    \caption{\textbf{Final Average Accuracy} $\bar{A}_F$ ($\uparrow$) and \textbf{Final Average Adjusted Forgetting} ($\bar{F}^*_F$) ($\downarrow$) on the \textbf{\classil} setting. \errino indicates a result obtained with rotation, \jayino a result obatined with $2\times2$ jigsaw puzzle.}
    \label{tab:cil_main}
\end{table*}

To better compare the effect of \methnam, we also include the results of a model jointly trained on all classes for one epoch (\textbf{Joint-online}) and for $30$ and $50$ epochs respectively on Seq.\ CIFAR-100 and Seq.\ \miniimagenet (\textbf{Joint-offline}). Also, we include the results of a model trained on the task sequence with no forgetting countermeasures (\textbf{Finetune}). 

\smallskip
\noindent\textbf{Architecture.}~We rely on ResNet18~\cite{he2016deep} as backbone in all experiments. For DualNet, we use this model as the slow learner and -- in line with~\cite{pham2021dualnet} -- construct the fast learner as a feed-forward network with the same number of convolutional layers as residual blocks in the slow learner.

Regardless of the underlying CL method, we define the feature extractor $g_\phi$ and the classification heads $h_\phi$ and $h_\xi$ by splitting the ResNet backbone at the second-last residual block; namely, $h_\phi$ and $h_\xi$ are comprised of the last residual block, followed by a linear projection onto the respective sets of classes $\mathcal{Y}=\cup_{i=1}^T\mathcal{Y}_i$ and $\mathcal{Y}_\mathcal{A}$. 

\noindent\textbf{Metrics.}~As a primary indicator of a model's performance at the end of OCL, we report its \textit{Final Average Accuracy} ($\bar{A}_F$). Let $a_i^j$ be the accuracy of the model at the end of task $j$ computed on the test set of task $\mathcal{T}_i$, $\bar{A}_F$ is computed as:
\begin{equation}
\bar{A}_F =
\frac{1}{T}\sum_{i=1}^{T}{a_i^T}.
\end{equation} 
To further assess learning as tasks progress, we report the \textit{Final Average Adjusted Forgetting} ($\bar{F}^*_F$), defined as follows:
\begin{equation}
  \begin{gathered}
    \bar{F}^*_F = \frac{1}{T-1} \sum_{i=1}^{T-1} \bigg[\frac{a_i^* - a_i^{T}}{a_i^*}\bigg]^+, \\
    \text{where}~~ a_i^* = \max_{t\in\{i,\dots,T-1\}}a_i^t,~~\forall i \in \{1,\dots,T-1\}.
  \end{gathered}
\end{equation}
$\bar{F}^*_F$ is a novel measure derived from the widely employed Forgetting metric~\cite{chaudhry2018riemannian} to facilitate the comparison between unevenly performing approaches. In particular, while the original Forgetting is upper-bounded by a model's accuracy, $\bar{F}^*_F$  varies in $[0, 100]$. $\bar{F}^*_F=100$ denotes a method that retains no accuracy on previous tasks (\textit{e.g.}, Finetune) and $\bar{F}^*_F=0$ one that has no performance decrease on past tasks. 

We repeat each experiment $10$ times and report the mean $\bar{A}_F$ and $\bar{F}^*_F$, and the standard deviation of the former. Please refer to the supplementary material for the standard deviations and statistical significance.
\subsection{Comparison with the State-Of-The-Art}
\label{sec:results}

\begin{table*}[t] 
    \centering
    \rowcolors{2}{}{lightgray}
    \begin{tabular}{lcccc}
    \toprule
    \textbf{\taskil} & \multicolumn{2}{c}{\textbf{Seq.\ CIFAR-100}} & \multicolumn{2}{c}{\textbf{Seq.\ \miniimagenet}} \\
    \midrule
    Joint-offline & \multicolumn{2}{c}{\result{82.69}{0.31}{-}}        & \multicolumn{2}{c}{\result{87.55}{0.16}{-}}  \\
    Joint-online  & \multicolumn{2}{c}{\result{54.12}{1.45}{-}}        & \multicolumn{2}{c}{\result{52.62}{0.98}{-}}         \\
    Finetune      & \multicolumn{2}{c}{\result{35.42}{3.11}{44.32}}    & \multicolumn{2}{c}{\result{31.55}{2.64}{28.75}}     \\
    \midrule
    \rowcolor{white}\textbf{Buffer Size} & 500 & 2000 & 2000 & 8000 \\
    \midrule
    ER-ACE~\cite{caccia2022new}      & \result{56.06}{1.41}{9.48} & \result{64.94}{1.11}{3.19} & \result{64.68}{1.21}{\underline{3.77}} & \result{66.17}{1.96}{\underline{4.10}} \\
    ~+~\methnam & \resultj{\textbf{61.60}}{1.78}{\underline{9.21}} & \resultj{\textbf{69.33}}{1.44}{\underline{3.04}} & \resultr{\textbf{68.02}}{1.32}{5.27} & \resultj{\textbf{69.13}}{1.42}{4.11} \\
    X-DER~\cite{boschini2022class}       & \result{63.10}{0.86}{4.31} & \result{69.00}{0.83}{1.38} & \result{67.67}{1.19}{4.71} & \result{68.97}{1.32}{4.39} \\
    ~+~\methnam & \resultj{\textbf{68.19}}{1.08}{\underline{2.98}} & \resultj{\textbf{73.45}}{0.55}{\underline{0.97}} & \resultj{\textbf{71.32}}{0.86}{\underline{3.01}} & \resultj{\textbf{72.39}}{0.89}{\underline{2.66}} \\
    CoPE~\cite{de2021continual}        & \result{51.89}{1.64}{23.46} & \result{66.56}{2.13}{7.48} & \result{70.10}{1.42}{\underline{4.89}} & \result{73.61}{2.36}{3.58} \\
    ~+~\methnam & \resultj{\textbf{60.19}}{1.23}{\underline{20.34}} & \resultj{\textbf{71.91}}{0.90}{\underline{6.42}} & \resultr{\textbf{71.17}}{1.05}{5.30} & \resultr{\textbf{75.33}}{1.27}{\underline{2.54}} \\
    \midrule
    DualNet~\cite{pham2021dualnet} & \result{49.38}{0.65}{25.20} & \result{57.05}{0.63}{13.85} & \result{68.43}{0.46}{9.99} & \result{73.89}{0.99}{5.54} \\
    ~+~\methnam & \resultr{\textbf{50.11}}{0.69}{\underline{23.94}} & \resultj{\textbf{59.66}}{0.39}{\underline{12.99}} & \resultr{\textbf{70.26}}{0.84}{\underline{7.39}} & \resultr{\textbf{76.97}}{0.33}{\underline{3.87}} \\
    \bottomrule
    \end{tabular}
    \caption{\textbf{Final Average Accuracy} $\bar{A}_F$ ($\uparrow$) and \textbf{Final Average Adjusted Forgetting} ($\bar{F}^*_F$) ($\downarrow$) on the \textbf{\taskil} setting. \errino indicates a result obtained with rotation, \jayino a result obtained with $2\times2$ jigsaw puzzle.}
    \label{tab:til_main}
\end{table*}
We include the results of our evaluation on Seq.\ CIFAR-100 and Seq.\ \miniimagenet for \classil and \taskil in Tab.~\ref{tab:cil_main} and~\ref{tab:til_main} respectively. For each experiment, we report the best performer among the 2$\times$2 jigsaw and rotation pretext tasks\footnote{Please refer to Sec.~\ref{sec:jigsaw} for a detailed comparison between the two choices of pretext task.}. The evidence we present strongly supports our initial claims, with \methnam improving the SOTA methods in all benchmarks. Specifically, we witness an improvement across the board regarding the $\bar{A}_F$, while $\bar{F}^*_F$ indicates stronger resistance against forgetting.

Interestingly, the effect of our regularization is maintained regardless of the choice of buffer size, with an average \classil improvement of $3.59$ and $3.40$ on Seq.\ CIFAR-100 and $3.12$ and $3.46$ on Seq.\ \miniimagenet. 
We find the only notable exception is in the case of DualNet on Seq.~CIFAR-100. Indeed, even without our regularization, the lower FAA and higher forgetting compared with the other baselines suggests that the model cannot profit from the memory buffer. This might be due to the fact that the slow learner is only trained with a CSSL objective on samples from the buffer, which limits the quality of its representation when the latter is of moderate size. However, its results on the challenging Seq.\ \miniimagenet, when combined with \methnam, suggest that such an effect can be mitigated by leveraging \textit{equivariant} SSL, which allows the fast learner to develop better representations during OCL.
\section{Model Analysis}
\begin{figure}
{
    \setlength{\tabcolsep}{0.1em}
    \centering
    \begin{tabular}{cc}
        \multicolumn{2}{c}{\includegraphics[width=\linewidth,trim={0 .8cm 0 .5cm},clip]{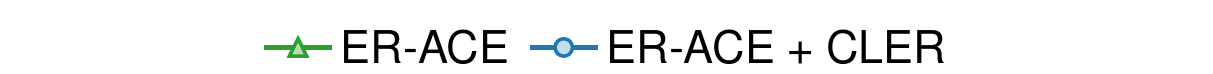}}\\
        \multicolumn{2}{c}{\includegraphics[width=\linewidth,trim={0 10.85cm 0 0cm},clip]{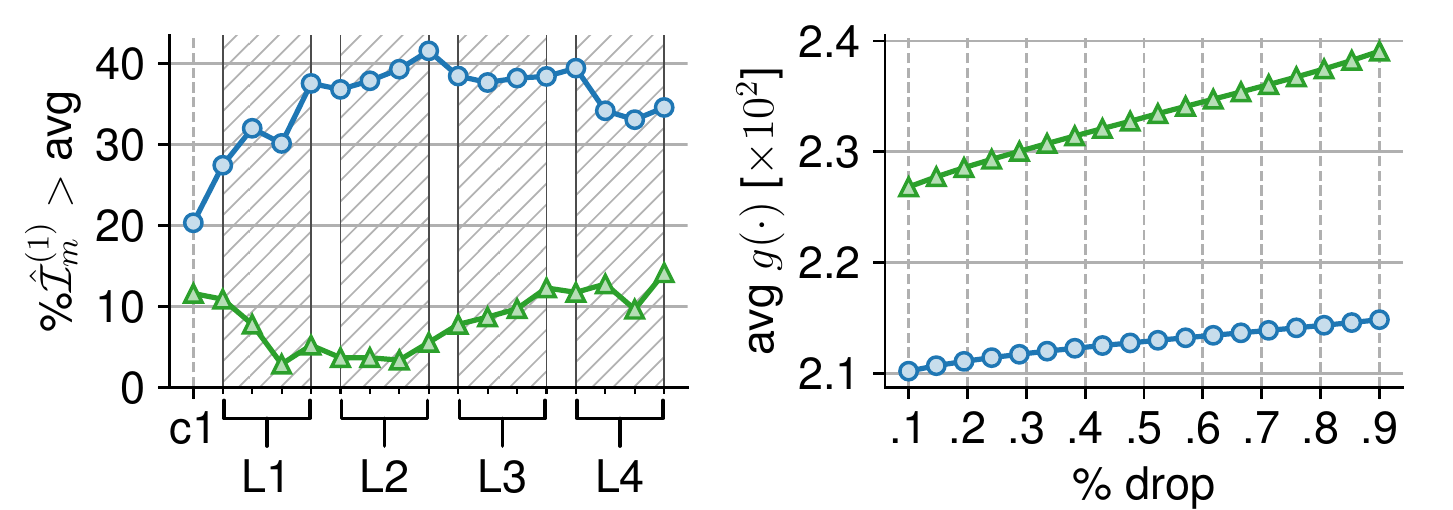}}\\
        \hspace{6.5em}(a) & \hspace{6.5em}(b) \\
        \multicolumn{2}{c}{\includegraphics[width=\linewidth,trim={0 6.6cm 0 5.35cm},clip]{content/figures/f3.pdf}}\\
        \multicolumn{2}{c}{\includegraphics[width=\linewidth,trim={0 .3cm 0 10.8cm},clip]{content/figures/f3.pdf}}\\
        \multicolumn{2}{c}{\hspace{2.em}(c)}\\
    \end{tabular}
    }
    \caption{\textbf{Structural analysis of ER-ACE with and without \methnam} on Seq.\ CIFAR-100. (a) Percentage of important neurons in each layer with \textbf{higher-than-average importance score} $\hat{\mathcal{I}}_m^{(1)}$; (b) within-layer \textbf{similarity score} $g$ after pruning with Geometric Median; (c) \textbf{accuracy after dropping} conv.\ filters and training on a few batches from $\mathcal{T}_6$, with the pre-drop accuracy serving as a target value (\textit{red} line) (\textit{best seen in colors}).}
    \label{fig:pruning}
\end{figure}
In the remainder, we analyze the various contributions of \methnam and gather further insights on its overall effect on the CL tasks. To the best of our knowledge, our work is the first to consider the effect of equivariant-based pretext tasks in an incremental setting.

\subsection{Effects of \methnam on the Backbone}
For an in-depth analysis of the effects induced on the backbone, we consider ER-ACE with and without \methnam and conduct three additional experiments, drawing inspiration from the Network Pruning literature~\cite{molchanov2019importance}. Our aim here is to unveil how the information carried by the learned features distributes across the parameters of the backbone. 

\noindent\textbf{Importance and redundancy.}~First, we quantify each parameter's contribution to the overall loss after training on Seq.\ CIFAR-100 by computing the \textit{importance measure} $\hat{\mathcal{I}}_m^{(1)}$ proposed in~\cite{molchanov2019importance}. 
In Fig.~\ref{fig:pruning}a, we focus on the convolutional layers and report the proportion of parameters whose importance score is higher than the layer's average to provide a compact per-layer evaluation.

Additionally, we perform a Geometric Median pruning~\cite{he2019filter} on the model, thus discarding those filters $\mathcal{F}_d$ that are the most redundant - \textit{i.e.}, averagely most similar to all others in the same layer. In Fig.~\ref{fig:pruning}b we report the average within-layer similarity $g$ for the discarded kernels:
\begin{equation}
g(\mathcal{F}_d)=\frac{1}{F}\sum_{j=1}^{F}|\mathcal{F}_d - \mathcal{F}_j|,
\end{equation}
with $F$ the total number of filters in the considered layer.

\begin{table*}[t]
    \centering   \rowcolors{2}{lightgray}{}
\begin{tabular}{lcccc}
\toprule
\textbf{Model} & \multicolumn{2}{c}{\textbf{Seq.\ CIFAR-100 (oCIL)}} & \multicolumn{2}{c}{\textbf{Seq.\ CIFAR-100 (oTIL)}} \\
\midrule 
\textbf{Buffer Size} & 500 & 2000 & 500 & 2000 \\

\midrule
ER-ACE~\cite{caccia2022new} & \result{20.17}{1.31}{38.75} & \result{26.95}{1.12}{23.69} & \result{56.06}{1.41}{9.48} & \result{64.94}{1.11}{3.19} \\
~+~CSSL & \result{20.89}{1.53}{36.03}              & \result{27.80}{0.15}{21.12}          & \result{56.22}{1.45}{9.88}          & \result{65.91}{0.20}{\underline{2.42}} \\
~+~\methnam & \resultj{\textbf{24.53}}{1.43}{\underline{33.76}} & \resultj{\textbf{30.89}}{1.54}{\underline{20.24}} & \resultj{\textbf{61.60}}{1.78}{\underline{9.21}} & \resultj{\textbf{69.33}}{1.44}{3.04} \\
\midrule
X-DER~\cite{boschini2022class} & \result{25.80}{0.98}{39.54} & \result{30.44}{0.80}{31.52} & \result{63.10}{0.86}{4.31} & \result{69.00}{0.83}{1.38} \\
~+~CSSL & \result{21.91}{1.08}{36.07}              & \result{23.59}{1.42}{40.53}          & \result{57.26}{3.14}{\underline{2.76}}          & \result{62.56}{1.85}{\underline{0.85}} \\
~+~\methnam & \resultj{\textbf{29.35}}{0.72}{\underline{35.56}} & \resultj{\textbf{34.57}}{0.96}{\underline{29.71}} & \resultj{\textbf{68.19}}{1.08}{2.98} & \resultj{\textbf{73.45}}{0.55}{0.97} \\
\midrule
CoPE~\cite{de2021continual} & \result{19.98}{1.49}{75.32} & \result{34.09}{2.12}{46.39} & \result{51.89}{1.64}{23.46} & \result{66.56}{2.13}{7.48} \\
~+~CSSL & \result{17.23}{0.62}{74.28}              & \result{25.76}{0.38}{54.72}          & \result{49.56}{0.40}{\underline{18.98}}          & \result{62.48}{0.24}{\underline{3.64}} \\
~+~\methnam & \resultj{\textbf{26.15}}{1.16}{\underline{69.28}} & \resultj{\textbf{38.48}}{1.48}{\underline{45.50}} & \resultj{\textbf{60.19}}{1.23}{20.34} & \resultj{\textbf{71.91}}{0.90}{6.42} \\
\bottomrule
\end{tabular}
%
    \caption{\textbf{Performance comparison} between our proposal \textbf{\methnam} and a similar \textbf{Contrastive-based SSL (CSSL)} method, as measured by \textbf{Final Average Accuracy} $\bar{A}_F\pm \operatorname{std}$ ($\uparrow$) and \textbf{Final Average Adjusted Forgetting} ($\bar{F}^*_F$) ($\downarrow$) on the Seq.\ CIFAR-100 benchmark.}
    \label{tab:abl_invariance}
    \vspace{-0.7em}
\end{table*}
Our results reveal that \methnam pushes the model to fit the learned task with dense configurations of parameters (higher $\hat{\mathcal{I}}_m^{(1)}$ in Fig.~\ref{fig:pruning}a) that are also more similar to each other (lower $g$ in Fig.~\ref{fig:pruning}b). We conjecture that this can be linked to the performance increase reported in Sec.~\ref{sec:results}: as the knowledge of a specific task does not rely on only a few parameters but instead appears more distributed, it is less likely that subsequent weights' updates will entirely erase the previously acquired knowledge. Moreover, the higher rate of important parameters, coupled with the higher redundancy, suggests that those important filters erased by forgetting could be restored as needed, by simply leveraging redundant groups of parameters.

\smallskip
\noindent\textbf{Recovery.}~To support our intuitions, we conducted an additional evaluation probing the dynamics of learning with \methnam. After training on the 6$^\text{th}$ task of Seq.\ CIFAR-100, we randomly drop a portion of the convolutional filters in our models and retrain using only the cross-entropy loss on a few batches from the same task, reporting the accuracy after each batch in Fig.~\ref{fig:pruning}c. Interestingly, the distributed importance induced by our training objective leads to a higher initial drop in accuracy for \methnam. However, our proposed approach swiftly recovers its performance, reaching the target pre-drop accuracy in fewer steps w.r.t.\ the baseline.
\subsection{Invariance \& Equivariance}
\label{sec:jigsaw}
While in previous sections we explored the role of equivariance as a regularizer for OCL, we now wish to better characterize the different pretext tasks, as well as compare with an invariance-based CSSL objective.
\begin{figure}
    \centering
    \includegraphics[width=\linewidth]{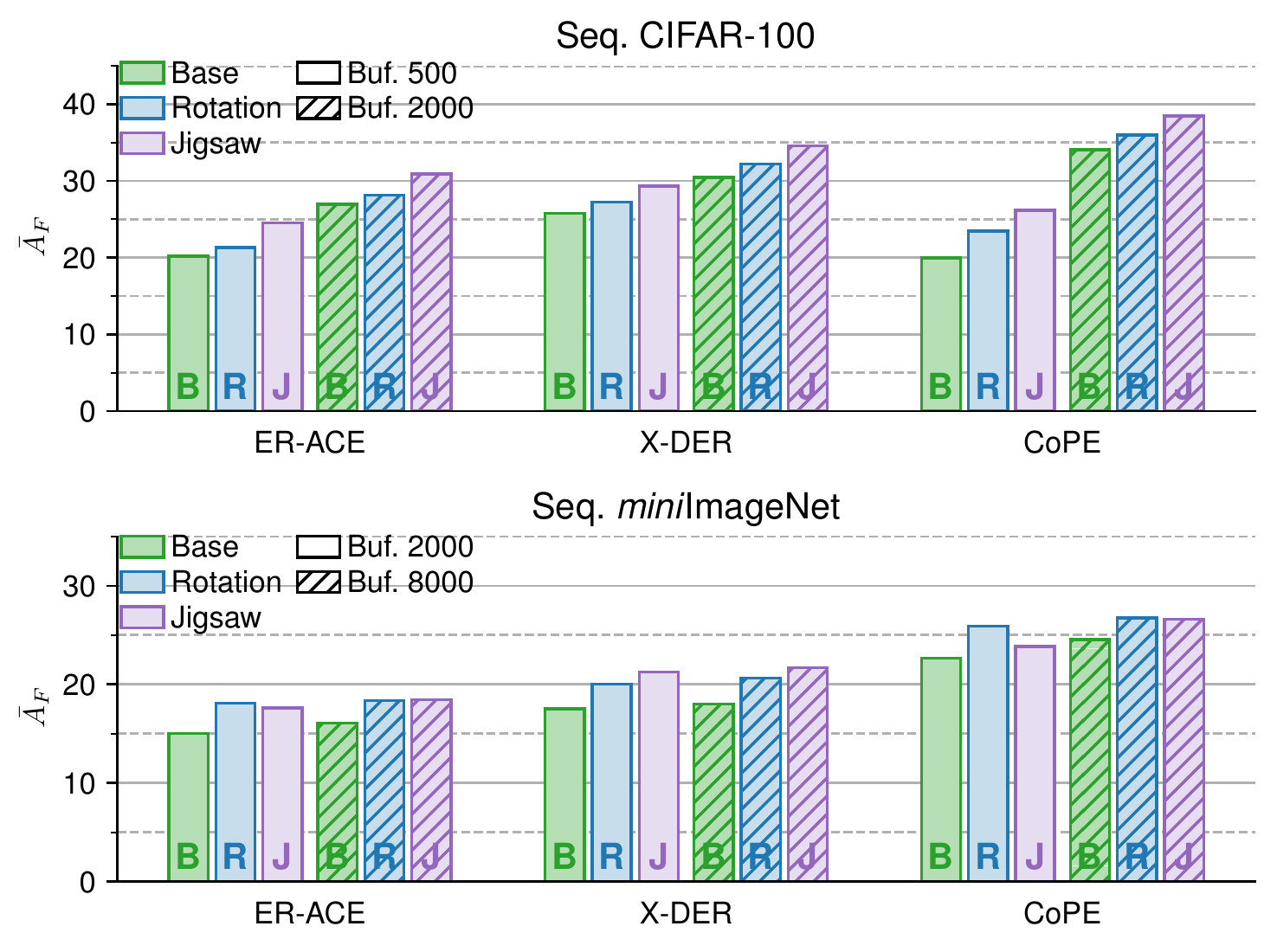}
    \caption{\textbf{Final Average Accuracy} $\bar{A}_F$ of various baseline methods when equipped with \textbf{different equivariant pretext tasks}: \textit{four-fold rotation prediction} and $2\times 2$ \textit{jigsaw solving}. Both methods achieve higher results w.r.t.\ the baseline, with jigsaw solving usually leading to the best performance (\textit{best seen in colors}).}
    \label{fig:jigsaw}
     \vspace{-0.5em}
\end{figure}

\noindent\textbf{Rotations \textit{vs} Jigsaw.}~The results presented so far depict a clear advantage of the jigsaw puzzle pretext task, which might suggest that the performance gain is not specifically tied to equivariance but to the former. To address such concern, in Fig.~\ref{fig:jigsaw} we present detailed results for the evaluation of Sec.~\ref{sec:results} on the \classil setting both with four-fold rotation and jigsaw puzzle.
Our results depict a clear advantage of both equivariant pretext tasks w.r.t.\ the baseline method. Moreover, the similar performance achieved by the two (especially on the challenging Seq.\ \miniimagenet benchmark) further proves our initial assumption about the effectiveness of equivariant-based SSL methods in CL.

\noindent\textbf{Comparison with CSSL methods.}~Our initial analysis shows that enforcing \textit{equivariance} to a set of input transformations efficiently allows \methnam to learn a representation robust against forgetting, by spreading the contribution of each feature on all the learnable parameters. This is in contrast with current CL literature, which instead relies on CSSL tasks~\cite{cha2021co2l,pham2021dualnet} to learn a representation that is \textit{invariant} to strong data augmentation and input transformations.

To further prove our contribution, in Tab.~\ref{tab:abl_invariance} we compare our proposal of an equivariant loss term against one that promotes invariance by means of a CSSL objective. For the latter, we take inspiration from~\cite{pham2021dualnet} and opt for Barlow Twins. Our results indicate a superior regularization effect for \methnam, with CSSL even hurting the performance in some scenarios. This suggests that the few training iterations allowed in OCL do not allow CSSL to transfer useful knowledge, thus eventually hindering incremental learning.
\begin{table}[t]
    \centering
    \rowcolors{2}{lightgray}{}
    \begin{tabular}{lccc}
    \toprule
       & \textbf{ER-ACE}~\cite{caccia2022new}              & \textbf{+~CSSL} & \textbf{+~\methnam}             \\
\midrule 
\textbf{Epochs}       &\multicolumn{3}{c}{\textbf{Buffer size} 500}\\
\midrule
1 (OCL)               & \result{20.17}{1.31}{38.75}                       & \result{20.89}{1.53}{36.03}                      & \resultj{\textbf{25.08}}{1.00}{\underline{32.84}} \\
5                     & \result{32.47}{1.76}{47.70}                       & \result{33.53}{0.59}{46.29}                      & \resultj{\textbf{34.88}}{0.22}{\underline{45.52}} \\
20                    & \result{37.38}{0.68}{\underline{46.79}}           & \result{37.78}{0.78}{50.55}                      & \resultj{\textbf{39.35}}{0.51}{46.84} \\
50                    & \result{37.94}{0.74}{51.49}                       & \result{39.61}{0.02}{\underline{43.75}}          & \resultj{\textbf{41.27}}{1.26}{46.78}             \\
\midrule
\rowcolor{white} \textbf{Epochs} &\multicolumn{3}{c}{\textbf{Buffer size} 2000}                                                                 \\
\midrule 
1 (OCL)               & \result{26.95}{1.12}{23.69}                       & \result{27.80}{0.15}{21.12}                      & \resultj{\textbf{30.89}}{1.54}{\underline{20.24}}\\
5                     & \result{42.35}{0.98}{27.49}                       & \result{43.62}{0.14}{27.11}                      & \resultj{\textbf{45.67}}{0.86}{\underline{24.92}}\\
20                    & \result{48.03}{0.70}{33.33}                       & \result{49.16}{0.29}{31.86}                      & \resultj{\textbf{50.27}}{0.81}{\underline{31.20}}\\
50                    & \result{49.05}{0.19}{33.91}                       & \result{50.66}{0.15}{34.48}                      & \resultj{\textbf{52.17}}{0.27}{\underline{32.56}}\\
\bottomrule
    \end{tabular}
    \vspace{0.17em}
    \caption{\textbf{Performance comparison} for \textbf{Equivariant-} and \textbf{Contrastive-based SSL} objectives in a \textbf{multi-epoch setting}, evaluated on Seq.\ CIFAR-100. We measure the \textbf{Final Average Accuracy} $\bar{A}_F$ ($\uparrow$) and find generally stronger performance for \methnam even when the online constraint is relaxed.}
    \vspace{-0.6em}
    \label{tab:ablat_epochs}
\end{table}

\noindent\textbf{Applicability to the multi-epoch setting.}~While we focus our evaluation on OCL, we reckon that our proposed approach might also prove beneficial in a less strict environment that allows for multiple iterations. Such a setting simulates a realistic low-latency scenario, where the desiderata is an algorithm capable of rapidly adapting to the changing data stream while retaining knowledge from the past. Results of this evaluation on the Seq.\ CIFAR-100 benchmark are summarized in Tab.~\ref{tab:ablat_epochs}. Due to space constraints, we only include results on the Class-Incremental scenario.

Unsurprisingly, as the number of epochs increases, the model can start to fully leverage the knowledge that comes from the stream. However, as CSSL tasks usually require a large number of iterations to converge, our \methnam remains a better choice for the task of preventing forgetting while boosting the representation of the base model.
\subsection{Is \methnam's advantage actually tied to OCL?}
\begin{table}[t]
  \centering
  \rowcolors{2}{}{lightgray}
  \begin{tabular}{lcc}
  \toprule
  \textbf{Method}          & \textbf{Seq.\ CIFAR-100} & \textbf{Seq.\ \miniimagenet} \\
  \midrule
   \textbf{Joint-offline}  & \resultVAR{69.85}{1.43} & \resultVAR{62.42}{1.13}     \\ 
   \rowcolor{white}
   ~+~CSSL                 & \resultVAR{70.24}{0.47} & \resultVAR{63.10}{0.61}     \\   
   \rowcolor{lightgray}
   ~+~\methnam             & \resultVAR{70.92\jayino}{0.74} & \resultVAR{63.11\jayino}{0.16}     \\ 
   \midrule
   \rowcolor{lightgray}
   \textbf{Joint-online}   & \resultVAR{23.14}{0.74} & \resultVAR{10.68}{0.67}     \\
    \rowcolor{white}
   ~+~CSSL                 & \resultVAR{23.16}{0.82} & \resultVAR{13.79}{0.79}     \\
   \rowcolor{lightgray}
   ~+~\methnam             & \resultVAR{28.38\jayino}{1.82} & \resultVAR{14.77\jayino}{0.78}     \\ 
   \bottomrule
  \end{tabular}
   \caption{\textbf{Accuracy} of \textbf{Joint} methods \textbf{with CSSL and \methnam}. The epochs are set to 30, 50 for CIFAR-100 and \textit{mini}Img respectively.}
   \label{tab:joints}
\end{table}
The consistently enhanced performance of baseline methods when combined with \methnam could raise the suspicion that SSL regularization is generally effective and not particularly relevant to Continual Learning \textit{per se}. To shed light on this point, we apply both CSSL and \methnam regularization on a multi-epoch Joint upper bound (Joint-offline) and report the results in Tab.~\ref{tab:joints}; this simple test clearly shows that -- if enough epochs are allowed and the method achieves full convergence -- the presence of additional SSL terms does not impact the attained accuracy significantly. 

To complement this result, we also apply the proposed technique on top of single-epoch Joint training. In this context, \methnam proves effective and more so than CSSL. In line with what shown in Fig.~\ref{fig:sgd-test}, this result confirms that SSL facilitates the convergence of the learner when having only few data-points and that the equivariant approach of \methnam is more sample-efficient than typical CSSL methods.

In conclusion, we summarize that \textbf{self-supervised regularization is not effective in a multi-epoch non-continual setting} (Tab.~\ref{tab:joints} \textit{top}); it becomes relevant in either single-epoch (Tab.~\ref{tab:joints} \textit{bottom}) or continual (Tab.~\ref{tab:ablat_epochs}) setting. Due to its enhanced sample efficiency, \textbf{the equivariant approach pursued by \methnam is particularly effective when fewer epochs are performed}. For this reason, its application is ideal for the OCL setting.

\subsection{Applicability to Data-Free Continual Learning}
\begin{table}[t]
  \centering
  \rowcolors{2}{}{lightgray}
  \begin{tabular}{lcc}
  \toprule
  \textbf{Method} & \textbf{Seq.\ CIFAR-100} & \textbf{Seq.\ \miniimagenet} \\
  \midrule
   LWF.MC~\cite{rebuffi2017icarl} & \result{36.15}{0.50}{49.78}                         & \result{20.75}{0.61}{63.67} \\ 
   ~+~\methnam                    & \resultr{\textbf{37.07}}{0.70}{\underline{49.37}}   & \resultr{\textbf{21.64}}{0.54}{\underline{62.79}} \\ 
   R-DFCIL~\cite{gao2022r}        & \result{34.98}{0.14}{54.59}                         & \result{13.15}{1.07}{83.47}       \\
   ~+~\methnam                    & \resultr{\textbf{36.74}}{0.34}{\underline{52.31}}   & \resultj{\textbf{18.80}}{0.90}{\underline{75.43}}  \\
   \bottomrule
  \end{tabular}
   \caption{Class-IL \textbf{Final Average Accuracy} $\bar{A}_F$ of \textbf{DFCIL} methods (\textit{no buffer}) \textbf{with and without \methnam}. We conduct 30, 50 epochs on CIFAR-100, \textit{mini}Img respectively.}
   \label{tab:dfcil}
\end{table}
The SOTA competitors on top of which we validate \methnam in Sec.~\ref{sec:exps} belong to the rehearsal-based family of CL methods. These represent by far the preferred approach in the challenging \classil scenario, on which the performance of other classes of methods is severely compromised~\cite{mai2022online,caccia2022new,gu2022not,zhang2022simple}.
However, a very recent line of works raises criticism on the adoption of replay, citing potential privacy issues~\cite{smith2021always,gao2022r}. They instead focus on the so-called \textbf{Data-Free Class-Incremental Learning (DFCIL)} setting, \textit{i.e.}, \textbf{multi-epoch} Class-Incremental Learning without a memory buffer.

To provide a clear picture of the flexibility of our proposal, we further showcase its application on top of two DFCIL methods: the model inversion-based Relation-Guided Representation Learning (R-DFCIL)~\cite{gao2022r} and the distillation-based Multi-Class Learning without Forgetting (LWF.MC)~\cite{rebuffi2017icarl}. The results in Tab.~\ref{tab:dfcil} illustrate that \methnam delivers a steady performance improvement even in DFCIL, which reveals that its effectiveness is not dependent on the availability of replay data.

\section{Conclusions}
We present \textbf{\methodname} (\textbf{\methnam}), a novel approach for \textit{Online Continual Learning} (OCL) that encourages representations to be sensitive to a set of input transformations. Our method introduces a regularization technique based on equivariant SSL pretext tasks (jigsaw puzzle solving and four-fold rotation prediction). By experimental means, we show that the application of \methnam to state-of-the-art methods consistently leads to better performance. Furthermore, we provide an in-depth analysis of the effect of \methnam on the parameters of the backbone network and compare it against other Contrastive Self-Supervised Learning methods. 

Our strong results with different choices of equivariant pretext tasks further support our initial hypothesis, laying the foundation for better OCL models based on equivariant constraints. We leave this analysis for future work.

{\small
\bibliographystyle{ieee_fullname}
\bibliography{biblio/compact,biblio/bibliography}
}

\end{document}


\title{On the Effectiveness of Equivariant Regularization\\for Robust Online Continual Learning -- Supplementary Material}

\maketitle


\section{Additional benchmark details}
\newcommand{\resultstd}[3]{$\pm$\ensuremath{#2}}
\subsection{Settings}
We provide further details regarding the experimental setting adopted for our comparison. In all experiments, we use the same batch size of $10$ for samples drawn from the stream and from the memory buffer. For the backbone, for Seq.\ CIFAR-100 we employ ResNet18~\cite{he2016deep} as in~\cite{buzzega2020dark,rebuffi2017icarl}, while for Seq.\ \miniimagenet we rely on a slim version of ResNet18 with $\nicefrac{1}{3}$ of the filters introduced in~\cite{lopez2017gradient}.

Regardless of the choice of the dataset, we train with SGD with \textit{RandomCrop} and \textit{RandomHorizontalFlip} as data augmentation.
\subsection{Baseline methods}
Differently from the other baselines considered, DualNet~\cite{pham2021dualnet} and CoPE~\cite{de2021continual} both include an inner iteration on the current batch. We fix the number of iterations for these methods to $5$ across all experiments. 
\begin{table}[t] 
    \centering
    \rowcolors{2}{}{lightgray}
    \begin{tabular}{lcccc}
    \toprule
    \textbf{\classil} & \multicolumn{2}{c}{\textbf{Seq.\ CIFAR-100}} & \multicolumn{2}{c}{\textbf{Seq.\ \miniimagenet}} \\
    \midrule    
    Joint-offline & \multicolumn{2}{c}{\resultstd{69.47}{0.20}{-}}                    & \multicolumn{2}{c}{\resultstd{63.31}{0.47}{-}} \\
    Joint-online & \multicolumn{2}{c}{\resultstd{23.14}{0.74}{-}}                    & \multicolumn{2}{c}{\resultstd{10.68}{0.67}{-}} \\
    Finetune        & \multicolumn{2}{c}{\resultstd{7.00}{0.15}{100}}    & \multicolumn{2}{c}{\resultstd{3.21}{0.10}{100}}          \\
    \midrule
\rowcolor{white}\textbf{Buffer Size} & 500 & 2000 & 2000 & 8000 \\
    \midrule
    ER-ACE~\cite{caccia2022new}      & \resultstd{20.17}{1.31}{38.75} & \resultstd{26.95}{1.12}{23.69} & \resultstd{15.03}{0.63}{35.01} & \resultstd{16.07}{1.29}{37.94} \\
    ~+~\methnam & \resultstd{\textbf{24.53}}{1.43}{\underline{33.76}} & \resultstd{\textbf{30.89}}{1.54}{\underline{20.24}} & \resultstd{\textbf{18.08}}{1.46}{\underline{32.53}} & \resultstd{\textbf{18.43}}{0.95}{\underline{33.22}} \\
    X-DER~\cite{boschini2022class}   & \resultstd{25.80}{0.98}{39.54} & \resultstd{30.44}{0.80}{31.52} & \resultstd{17.51}{1.08}{34.25} & \resultstd{18.01}{1.54}{50.84} \\
    ~+~\methnam & \resultstd{\textbf{29.35}}{0.72}{\underline{35.56}} & \resultstd{\textbf{34.57}}{0.96}{\underline{29.71}} & \resultstd{\textbf{21.26}}{0.91}{\underline{34.07}} & \resultstd{\textbf{21.71}}{1.30}{\underline{34.76}} \\
    CoPE~\cite{de2021continual}      & \resultstd{19.98}{1.49}{75.32} & \resultstd{34.09}{2.12}{46.39} & \resultstd{22.67}{1.32}{57.96} & \resultstd{24.54}{2.67}{55.09} \\
    ~+~\methnam & \resultstd{\textbf{26.15}}{1.16}{\underline{69.28}} & \resultstd{\textbf{38.48}}{1.48}{\underline{45.50}} & \resultstd{\textbf{25.91}}{1.67}{\underline{57.73}} & \resultstd{\textbf{26.76}}{1.17}{\underline{52.69}} \\
    \midrule
    DualNet~\cite{pham2021dualnet}   & \resultstd{11.09}{0.51}{92.42} & \resultstd{19.93}{0.74}{73.44} & \resultstd{16.21}{1.13}{80.35} & \resultstd{25.33}{1.47}{59.60} \\
    ~+~\methnam & \resultstd{\textbf{11.89}}{0.61}{\underline{89.97}} & \resultstd{\textbf{20.88}}{0.90}{\underline{73.02}} & \resultstd{\textbf{18.66}}{1.00}{\underline{72.74}} & \resultstd{\textbf{30.90}}{0.86}{\underline{52.14}} \\
    \bottomrule
    \end{tabular}
    \caption{\textbf{Standard deviation} of our main experiments on the \textbf{oCIL} setting (Tab.~\ref{tab:cil_main} of the main manuscript).}
    \label{tab:cil_std}
\end{table}
\subsection{Statistical significance of our evaluation}
As mentioned in the manuscript, we repeat each experiment $10$ times and average the Final Average Accuracy ($\bar{A}_F$) for a solid comparison. Here, we present the standard deviation (Tab.~\ref{tab:cil_std}) and the statistical significance (Fig.~\ref{tab:cil_sign}) of each result. The latter measures the probability that the performance gain of \methnam is due to randomness rather than on the model itself. Specifically, we compute it as the $p$-value of a Student's $t$-test. Values lower than $5\%$ indicate a statistically solid performance gain. In our case, as shown in~\ref{tab:cil_sign} the $p$-values obtained indicate solid contribution as most of our experiments have a $p$-value even lower than $1\%$.
\begin{table*}[t]
    \centering
    \rowcolors{2}{}{lightgray}
    \begin{tabular}{lcccc}
    \toprule
    \textbf{\classil} & \multicolumn{2}{c}{\textbf{Seq.\ CIFAR-100}} & \multicolumn{2}{c}{\textbf{Seq.\ \miniimagenet}} \\
    \midrule
\rowcolor{white}\textbf{Buffer Size} & 500 & 2000 & 2000 & 8000 \\
    \midrule
    ER-ACE~\cite{caccia2022new} & \plusresult{4.36}{0.00001} & \plusresult{3.94}{0.00000} & \plusresult{3.05}{0.00001} & \plusresult{2.36}{0.00020} \\
    X-DER~\cite{boschini2022class} & \plusresult{3.55}{0.00000} & \plusresult{4.13}{0.00000} & \plusresult{3.75}{0.00000} & \plusresult{3.70}{0.00002} \\
    CoPE~\cite{de2021continual} & \plusresult{6.17}{0.00000} & \plusresult{4.39}{0.00004} & \plusresult{3.24}{0.00014} & \plusresult{2.22}{0.02697} \\
    DualNet~\cite{pham2021dualnet} & \plusresult{0.80}{0.00517} & \plusresult{0.95}{0.01894} & \plusresult{2.45}{0.00007} & \plusresult{5.57}{0.00000} \\
    \bottomrule
    \end{tabular}
    \caption{\textbf{Accuracy gain} of \methnam \textbf{over the baseline} and its relative $p$-value within square brackets.}
    \label{tab:cil_sign}
\end{table*}
\subsection{Code implementation}
We perform experiments following the Mammoth framework~\cite{buzzega2020dark,boschini2022class}. Our code will be released and made publicly available upon acceptance.
\section{Chosen hypterparameters}
\subsection{Seq.\ CIFAR-100}%
\begin{scriptsize}
\noindent\begin{small}\textbf{No buffer}\end{small}\\
\textbf{Finetune}: $lr$:0.01\\
\textbf{Joint-online (1 epoch)}: $lr$:0.01\\
\textbf{Joint-offline (30 epochs)}: $lr$:0.01\\

\noindent\begin{small}\textbf{Buffer size: 500}\end{small}\\
\textbf{ER-ACE}: $lr$:0.01\\
\textbf{ER-ACE + CSSL}: $lr$:0.01;~$\lambda_r$:0.0001\\
\textbf{ER-ACE + \methnam\jayino}: $lr$:0.01;~$\lambda_r$:1.5\\
\textbf{X-DER}: $lr$:0.01;~$\beta$:0.8;~$\alpha$:0.8;~$m$:0.3;~$\eta$:0.1;~$\gamma$:0.85\\
\textbf{X-DER + CSSL}: $lr$:0.03;~$\beta$:0.8;~$\alpha$:0.5;~$\lambda_r$:0.01;~$m$:0.3;~$\eta$:0.1;~$\gamma$:0.85\\
\textbf{X-DER + \methnam\jayino}: $lr$:0.01;~$\beta$:0.8;~$\alpha$:0.8;~$\lambda_r$:0.8;~$m$:0.3;~$\eta$:0.1;~$\gamma$:0.85\\
\textbf{CoPE}: $lr$:0.005;~$\tau$:0.05;~$\alpha$:0.99\\
\textbf{CoPE + CSSL}: $lr$:0.005;~$\lambda_r$:0.05;~$\tau$:0.05;~$\alpha$:0.7\\
\textbf{CoPE + \methnam\jayino}: $lr$:0.001;~$\lambda_r$:1.5;~$\tau$:0.05;~$\alpha$:0.99\\
\textbf{DualNet}: $lr$:0.03;~$\tau$:2.0;~$\lambda_{tr}$:15.0;~$\beta$:0.05;~$slownet-lr$:0.005\\
\textbf{DualNet + \methnam\errino}: $lr$:0.01;~$\lambda_r$:0.03;~$\tau$:2.0;~$\lambda_{tr}$:13.0;~$\beta$:0.05;~$slownet-lr$:0.03\\

\noindent\begin{small}\textbf{Buffer size: 2000}\end{small}\\
\textbf{ER-ACE}: $lr$:0.01\\
\textbf{ER-ACE + CSSL}: $lr$:0.01;~$\lambda_r$:0.0001\\
\textbf{ER-ACE + \methnam\jayino}: $lr$:0.01;~$\lambda_r$:1.5\\
\textbf{X-DER}: $lr$:0.01;~$\beta$:0.8;~$\alpha$:0.8;~$m$:0.3;~$\eta$:0.1;~$\gamma$:0.85\\
\textbf{X-DER + CSSL}: $lr$:0.03;~$\beta$:0.8;~$\alpha$:0.8;~$\lambda_r$:0.01;~$m$:0.3;~$\eta$:0.1;~$\gamma$:0.85\\
\textbf{X-DER + \methnam\jayino}: $lr$:0.01;~$\beta$:0.8;~$\alpha$:0.8;~$\lambda_r$:0.8;~$m$:0.3;~$\eta$:0.1;~$\gamma$:0.85\\
\textbf{CoPE}: $lr$:0.005;~$\tau$:0.05;~$\alpha$:0.99\\
\textbf{CoPE + CSSL}: $lr$:0.005;~$\lambda_r$:0.05;~$\tau$:0.05;~$\alpha$:0.7\\
\textbf{CoPE + \methnam\jayino}: $lr$:0.001;~$\lambda_r$:1.0;~$\tau$:0.05;~$\alpha$:0.99\\
\textbf{DualNet}: $lr$:0.03;~$\tau$:2.0;~$\lambda_{tr}$:15.0;~$\beta$:0.05;~$slownet-lr$:0.03\\
\textbf{DualNet + \methnam\jayino}: $lr$:0.03;~$\lambda_r$:1.0;~$\tau$:2.0;~$\lambda_{tr}$:15.0;~$\beta$:0.05;~$slownet-lr$:0.03\\
\end{scriptsize}
\subsection{Seq.\ \textbf{\textit{mini}}ImageNet}
\begin{scriptsize}
\noindent\begin{small}\textbf{No buffer}\end{small}\\
\textbf{Finetune}: $lr$:0.03\\
\textbf{Joint-online (1 epoch)}: $lr$:0.01\\
\textbf{Joint-offline (50 epochs)}: $lr$:0.03\\

\noindent\begin{small}\textbf{Buffer size: 2000}\end{small}\\
\textbf{ER-ACE}: $lr$:0.1\\
\textbf{ER-ACE + \methnam\errino}: $lr$:0.03;~$\lambda_r$:0.3\\
\textbf{X-DER}: $lr$:0.03;~$m$:0.3;~$\eta$:0.1;~$\beta$:1.5;~$\alpha$:0.3;~$\gamma$:0.85\\
\textbf{X-DER + \methnam\jayino}: $lr$:0.03;~$\lambda_r$:0.5;~$m$:0.3;~$\eta$:0.1;~$\beta$:1.0;~$\alpha$:0.3;~$\gamma$:0.85\\
\textbf{CoPE}: $lr$:0.005;~$\tau$:0.05;~$\alpha$:0.7\\
\textbf{CoPE + \methnam\errino}: $lr$:0.001;~$\tau$:0.05;~$\lambda_r$:0.1;~$\alpha$:0.6\\
\textbf{DualNet}: $lr$:0.03;~$\tau$:2.0;~$\lambda_{tr}$:10.0;~$\beta$:0.05;~$slownet-lr$:0.03\\
\textbf{DualNet + \methnam\errino}: $lr$:0.03;~$\lambda_r$:1.5;~$\tau$:2.0;~$\lambda_{tr}$:13.0;~$\beta$:0.05;~$slownet-lr$:0.01\\

\noindent\begin{small}\textbf{Buffer size: 8000}\end{small}\\
\textbf{ER-ACE}: $lr$:0.1\\
\textbf{ER-ACE + \methnam\jayino}: $lr$:0.03;~$\lambda_r$:1.0\\
\textbf{X-DER}: $lr$:0.01;~$m$:0.3;~$\eta$:0.1;~$\beta$:0.8;~$\alpha$:0.8;~$\gamma$:0.85\\
\textbf{X-DER + \methnam\jayino}: $lr$:0.03;~$\lambda_r$:0.5;~$m$:0.3;~$\eta$:0.1;~$\beta$:1.0;~$\alpha$:0.3;~$\gamma$:0.85\\
\textbf{CoPE}: $lr$:0.005;~$\tau$:0.05;~$\alpha$:0.7\\
\textbf{CoPE + \methnam\errino}: $lr$:0.0005;~$\tau$:0.05;~$\lambda_r$:0.1;~$\alpha$:0.7\\
\textbf{DualNet}: $lr$:0.1;~$\tau$:2.0;~$\lambda_{tr}$:10.0;~$\beta$:0.05;~$slownet-lr$:0.03\\
\textbf{DualNet + \methnam\errino}: $lr$:0.01;~$\lambda_r$:0.3;~$\tau$:2.0;~$\lambda_{tr}$:7.0;~$\beta$:0.05;~$slownet-lr$:0.03\\
\end{scriptsize}
%

{\small
\bibliographystyle{ieee_fullname}
\bibliography{biblio/compact,biblio/bibliography}
}